\documentclass[letterpaper, 10pt, conference, twoside]{ieeeconf}
\makeatletter
\let\NAT@parse\undefined
\makeatother
\usepackage{graphicx}
\usepackage[utf8]{inputenc}
\usepackage{authblk}
\usepackage{url}
\usepackage{amsfonts}
\usepackage{amsmath}
\usepackage{balance}
\usepackage{cite}
\usepackage{float}
\usepackage{makecell}
\usepackage{booktabs}
\usepackage[table]{xcolor}
\usepackage{tabularx}     
\usepackage{cuted}
\usepackage{subfig}
\usepackage{placeins}
\usepackage{dblfloatfix}
\usepackage{titlesec}
\let\labelindent\relax
\usepackage{enumitem}
\usepackage{hyperref}
\titlespacing{\section}{0pt}{*1.3}{*0.3}
\titlespacing{\subsection}{0pt}{*0.3}{*0.3}
\newcommand{\ours}{Phys2Real}

\usepackage[most]{tcolorbox}
\definecolor{StanfordCardinal}{HTML}{9F1C1F} 
\definecolor{StanfordLightRed}{HTML}{F4E6E7} 
\definecolor{StanfordMidRed}{HTML}{D8A6A7}   

\tcbset{
  stanfordstyle/.style={
    colback=StanfordLightRed,      
    colframe=StanfordCardinal,     
    colbacktitle=StanfordMidRed,   
    coltitle=black,                
    boxrule=1pt,                   
    arc=3pt,
    left=6pt,
    right=6pt,
    top=5pt,
    bottom=5pt,
    enhanced,
    fonttitle=\bfseries,
  },
}

\begin{document}

\title{\fontsize{19pt}{24pt}\selectfont
Phys2Real: Fusing VLM Priors with Interactive Online Adaptation for Uncertainty-Aware Sim-to-Real Manipulation
\thanks{This work is in part supported by ONR N00014-23-1-2355, ONR MURI N00014-22-1-2740, ONR MURI N00014-24-1-2748, NSF RI \#2338203, and NSF FRR grant 2342246. M. Wang is supported by the NASA NSTGRO Fellowship and NSF grant 2342246. S. Tian and A. Swann are supported by NSF GRFP Grant No. DGE-1656518 and DGE-2146755, respectively.}
} 

\IEEEoverridecommandlockouts
\overrideIEEEmargins

\author{Maggie Wang\textsuperscript{1}, Stephen Tian\textsuperscript{1}, Aiden Swann\textsuperscript{1}, Ola Shorinwa\textsuperscript{2}, Jiajun Wu\textsuperscript{1}, and Mac Schwager\textsuperscript{1}%
\thanks{\textsuperscript{1}Stanford University, Stanford, CA, USA.}
\thanks{\textsuperscript{2}Princeton University, Princeton, NJ, USA.}
}

\maketitle
\setlength{\abovedisplayskip}{2pt}
\setlength{\belowdisplayskip}{2pt}
\setlength{\abovedisplayshortskip}{2pt}
\setlength{\belowdisplayshortskip}{2pt}
\setlength{\dbltextfloatsep}{2pt plus 1pt minus 1pt}
\setlength{\intextsep}{4pt}
\setlength{\floatsep}{4pt}

\titlespacing{\section}{0pt}{*0.8}{*0.2}

\begin{strip}
\vspace{-21mm}
\centering
\includegraphics[width=\textwidth]{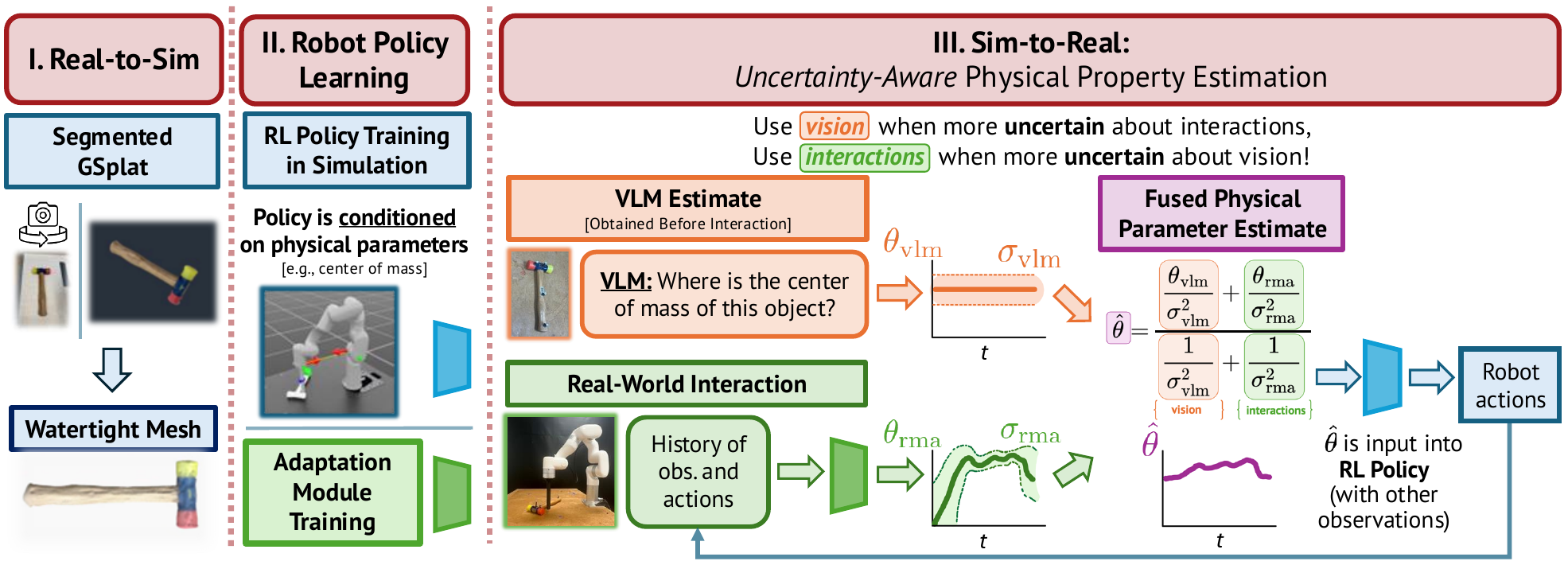}

\captionof{figure}{\small Phys2Real is a real-to-sim-to-real pipeline for robotic manipulation that combines VLM-based physical parameter estimation with interaction-based adaptation through uncertainty-aware fusion. It comprises three stages:
(I) \textit{real-to-sim}: simulation-ready mesh reconstruction from segmented Gaussian Splats,
(II) \textit{policy learning}: reinforcement learning conditioned on physical parameters, e.g., center of mass, and 
(III) \textit{sim-to-real transfer}: uncertainty-aware fusion of VLM priors and interaction-based estimates for online adaptation.}
\label{pipeline}
\vspace{-1.2em}
\end{strip}

\begin{abstract}
Learning robotic manipulation policies directly in the real world can be expensive and time-consuming. While reinforcement learning (RL) policies trained in simulation present a scalable alternative, effective sim-to-real transfer remains challenging, particularly for tasks that require precise dynamics. To address this, we propose Phys2Real, a real-to-sim-to-real RL pipeline that combines vision-language model (VLM)-inferred physical parameter estimates with interactive adaptation through uncertainty-aware fusion. Our approach consists of three core components: (1) high-fidelity geometric reconstruction with 3D Gaussian splatting, (2) VLM-inferred prior distributions over physical parameters, and (3) online physical parameter estimation from interaction data. Phys2Real conditions policies on interpretable physical parameters, refining VLM predictions with online estimates via ensemble-based uncertainty quantification. On planar pushing tasks of a \mbox{T-block} with varying center of mass and a hammer with an off-center mass distribution, Phys2Real achieves substantial improvements over a domain randomization baseline: 100\% vs 79\% success rate for the bottom-weighted T-block, 57\% vs 23\% in the challenging top-weighted T-block, and 15\% faster average task completion for hammer pushing. Ablation studies indicate that the combination of VLM and interaction information is essential for success. \textbf{Project website:} \url{https://phys2real.github.io/}
\end{abstract}



\IEEEpeerreviewmaketitle

\titlespacing{\section}{0pt}{*0.8}{*0.2}


\section{Introduction}

Deploying robotic manipulation policies trained in simulation to the real world remains a fundamental challenge, especially for tasks requiring fine-grained physical dynamics. Robots must adapt to varying object properties such as friction, mass distribution, and compliance, which significantly affect manipulation outcomes but are difficult to model precisely. While learning from demonstrations has shown significant promise, it often lacks the physical grounding and reasoning needed to adapt to novel objects. Reinforcement learning (RL) provides a mechanism for real-time adaptation, but bridging the sim-to-real gap remains a critical obstacle. 

Domain randomization (DR) is the dominant approach for sim-to-real transfer when training robotic policies with RL. By training policies across randomized simulation parameters, DR trains policies robust to real-world variations~\cite{tobin_domain_2017}, but they may generalize poorly to out-of-distribution object physical properties. Even when dynamics lie within the training distribution, policies often default to averaged behaviors that may not account for object-specific variations. 

We propose an alternative approach that efficiently balances robustness with performance. Specifically, we train policies to be not only \emph{robust} to broad parameter ranges, but also \emph{adaptive} to the specific physical properties of objects for superior performance, motivated by the question: \textbf{Can combining visual physical reasoning with interactive learning improve robot manipulation performance in real-world environments after training in simulation?}
 


Humans demonstrate remarkable exploration behaviors when manipulating novel objects in unseen settings. Initial judgments are formed about an object's physical properties from visual appearance, and then these estimates are refined through interaction \cite{Ernst2002}. This integration of perception and physical reasoning enables humans to adapt their manipulation strategies to new instances of objects with potentially varying properties without extensive interaction. Inspired by this observation, our approach seeks to provide robots with a similar ability to estimate and adapt to physical properties. 

We present \textbf{Phys2Real}, a framework that bridges the sim-to-real gap by combining three components: 
\begin{enumerate}[leftmargin=*]
    \item \textbf{Uncertainty-aware fusion of VLM priors with interactive adaptation}: We demonstrate that VLMs can provide physical parameter estimates (e.g., center of mass) that improve manipulation performance when combined with interaction-based parameter estimation. Our work demonstrates the novel application of VLMs to physical parameter estimation for real-time low-level closed-loop control, beyond standard high-level planning.
    \item \textbf{Ensemble-based uncertainty quantification}: We decompose uncertainty into epistemic and aleatoric components for interaction-based parameter estimation, and then combine these estimates with VLM priors using inverse-variance weighting. This addresses the limitations of existing adaptation methods that cannot incorporate external priors and enables adaptation during intermittent contact scenarios common in manipulation.
    \item \textbf{Physically-informed digital twins}: We combine 3D Gaussian Splatting reconstructions with online physical property estimation to create digital twins that incorporate both geometric and physical information, enabling strong sim-to-real transfer compared to purely visual or adaptation-based approaches.
\end{enumerate}

Our approach builds on rapid motor adaptation (RMA) \cite{kumar_rma_2021} but adapts it for physically interpretable parameter estimation rather than learned latents, enabling direct combination with VLM priors through uncertainty weighting.
%
%
The key insight is that policies can condition directly on physically interpretable parameters estimated from vision and refined through interaction, rather than learning averaged behaviors across parameter distributions as in DR. Unlike recent work that uses VLM estimates to initialize simulation parameters \cite{elhafsi2025scan}, our method updates these estimates online using real-world interaction histories during closed-loop control.

We evaluate our approach on two planar pushing tasks: (1)~\emph{T-block pushing} with varying center of mass (CoM) achieved by placing a small metal weight at different locations of a 3D printed T-block, and (2)~\emph{Hammer pushing} with an off-center mass distribution. Our results demonstrate consistent improvements: Phys2Real achieves a success rate of 100\% vs. 79\% for DR when the weight is at the bottom of the T-block, and 57.14\% vs. 23\% for DR in the more challenging configuration with weight at the top of the T-block. Furthermore, on the hammer pushing task, it achieves 15\% faster average task completion. 

This work demonstrates that VLMs can provide interpretable, uncertainty-calibrated physical estimates for manipulation that can be combined with interaction history. Phys2Real is a step towards more general, adaptive robotic systems that learn from perception and physical interaction.



\section{Related Work}
Our approach bridges sim-to-real transfer methods, policy adaptation techniques, digital twin reconstruction, and physical reasoning with foundation models. While prior work has explored these areas individually, Phys2Real uniquely combines high-fidelity 3D reconstructions with VLM-based physical parameter estimates and online adaptation to learn adaptive policies for robotic manipulation.

\subsection{Domain Randomization and System Identification} 
The sim-to-real gap remains a fundamental challenge for deploying policies learned in simulation to real-world environments. One class of approaches performs domain randomization (DR)~\cite{tobin_domain_2017, peng2018simtoreal} to address this, randomizing simulation dynamics during training to develop policies robust to a range of environmental variations. 
Although DR-trained policies have shown success in several domains~\cite{Sadeghi-RSS-17, openai_solving_2019, exarchos_policy_2021, chen_understanding_2022}, they often default to averaged behaviors that sacrifice performance for robustness, failing to adapt to object-specific variations. Our method trains simulation parameter-conditioned policies that adapt online to specific conditions, rather than an unconditioned policy trained to be robust across all conditions.


An alternative approach is to perform system identification \cite{10.5555/21413} to explicitly calibrate simulation parameters to match real-world observations. However, these methods often require manual parameter tuning and yield static models that cannot adapt to varying deployment conditions. We build on their extensions to online policy adaptation methods.

\subsection{Online Policy Adaptation} 
Online policy adaptation methods train universal policies conditioned on environment parameters~\cite{yu2017preparing} and perform online system identification for inference-time adaptation~\cite{xu2019densephysnet}.
Rapid Motor Adaptation (RMA)~\cite{kumar_rma_2021}, initially demonstrated for legged locomotion, trains an RL policy with an adaptation model that uses privileged information during simulation training and infers environment properties from interaction history at runtime. 
Related techniques have shown success across drone flight~\cite{low2025sous}, prehensile manipulation~\cite{liang_rapid_2024}, tool use~\cite{zhang2025dynamics}, and dexterous in-hand manipulation~\cite{qi2022hand}.

Although RMA is effective in scenarios such as locomotion and in-hand manipulation, where the robot makes frequent contact with its environment and manipulated objects, it is challenging to deploy to general manipulation settings due to intermittent contacts that lead to uninformative histories. Our work addresses this challenge by combining VLM estimates with an ensemble of online adaptation models that provide uncertainty-aware predictions. To do this, \ours{} conditions policies directly on physically interpretable parameters (e.g., center of mass), rather than learned latents. 


\subsection{Digital Twin Simulations and Photorealistic Rendering}
Another set of methods attempts to train transferable policies in simulation by creating photorealistic simulation environments, minimizing the sim-to-real gap.
Neural Radiance Fields (NeRF) \cite{mildenhall2021nerf} and Gaussian Splatting (GSplat) \cite{kerbl20233dgaussiansplattingrealtime, qureshi_splatsim_2024, li_robogsim_2024, wu2025rlgsbridge3dgaussiansplatting} can reconstruct high-fidelity 3D scenes from a series of images. They can be used for simulated policy training~\cite{qureshi_splatsim_2024}, synthetic data generation~\cite{yu2025real2render2realscalingrobotdata}, or as a mirrored surrogate environment~\cite{abou2025real}.

Existing digital twins focus on visual fidelity, neglecting object physical properties. They usually rely on conventional physics engines with default parameters that may not match real-world dynamics. 
Towards addressing this, some approaches allow manual physical property specification~\cite{torne_reconciling_2024} or estimate object geometry or robot physical properties from interaction data~\cite{pfaff2025_scalable_real2sim, moran2025splatting, wang2025embodiedreamer, bianchini2025vysics, li_pin-wm_2025}. 
Most relevant to our work, Elhafsi et al.~\cite{elhafsi2025scan} use VLM estimates to initialize simulation parameters for simulated objects. However, our method updates these initial estimates online using real-world interaction histories while executing in closed-loop.


\subsection{VLMs for Physical Reasoning} 

Recent work has demonstrated the physical reasoning capabilities of vision-language models (VLMs). 
PhysObjects~\cite{gao_physically_2024} fine-tunes InstructBLIP to estimate attributes such as material properties, weight, and fragility from images. However, this work focuses primarily on using these estimates for high-level planning rather than low-level control.

Phys2Real builds on these insights by using VLMs to estimate physical parameters from images. Unlike previous work that uses VLMs primarily for high-level planning~\cite{ning2025prompting} or affordance prediction~\cite{guo2024phygrasp}, we directly incorporate VLM-estimated physical parameters into the control policy, enabling more accurate manipulation.



\section{Real-to-Sim-to-Real Policy Learning with Uncertainty-Aware Adaptation}

Our method, Phys2Real, addresses the challenge of sim-to-real transfer by creating physically-informed digital twins and training adaptive manipulation policies. As illustrated in Fig.~\ref{pipeline}, our approach consists of three stages: (1) \textbf{real-to-sim} reconstruction to create geometrically accurate simulation assets, (2) \textbf{physics-parameter conditioned policy learning} in simulation with uncertainty-aware adaptation, and (3)~\mbox{\textbf{sim-to-real transfer}} that fuses VLM priors with estimates inferred from interaction data. 

\begin{figure}[!t]
\centering
\includegraphics[width=0.5\textwidth]{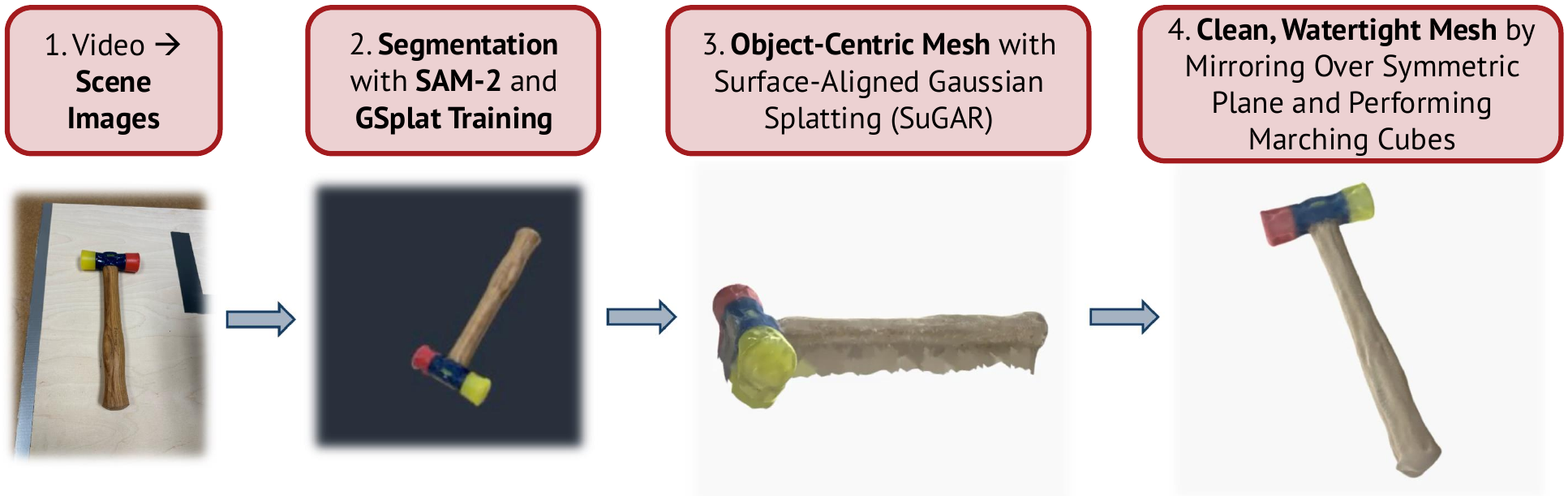}
\caption{\small \textbf{Real-to-sim mesh reconstruction pipeline.} From a video of the object, we extract image frames and segment the target object using SAM-2. We then train a GSplat and extract a surface-aligned object-centric mesh using SuGaR~\cite{guedon_sugar_2023}. Finally, we generate a clean, watertight mesh, resulting in a simulation-ready asset.}
\label{mesh-reconstruction}
\end{figure}

\subsection{Real-to-Sim Scene Reconstruction}

For objects without known meshes, we must first obtain simulation-ready assets, ideally with minimal manual intervention. Thus, we develop a pipeline to reconstruct meshes directly from video frames. As shown in Fig.~\ref{mesh-reconstruction}, we begin by capturing images of the target object and then segmenting them using SAM-2 \cite{ravi_sam_2024}. We then train a 3D GSplat~\cite{kerbl20233dgaussiansplattingrealtime} on the object foreground images and extract a watertight mesh using SuGaR~\cite{guedon_sugar_2023}. This process generates geometrically accurate assets for our simulation environment.

\begin{figure}[!tb]
\centering
\includegraphics[width=0.5\textwidth]{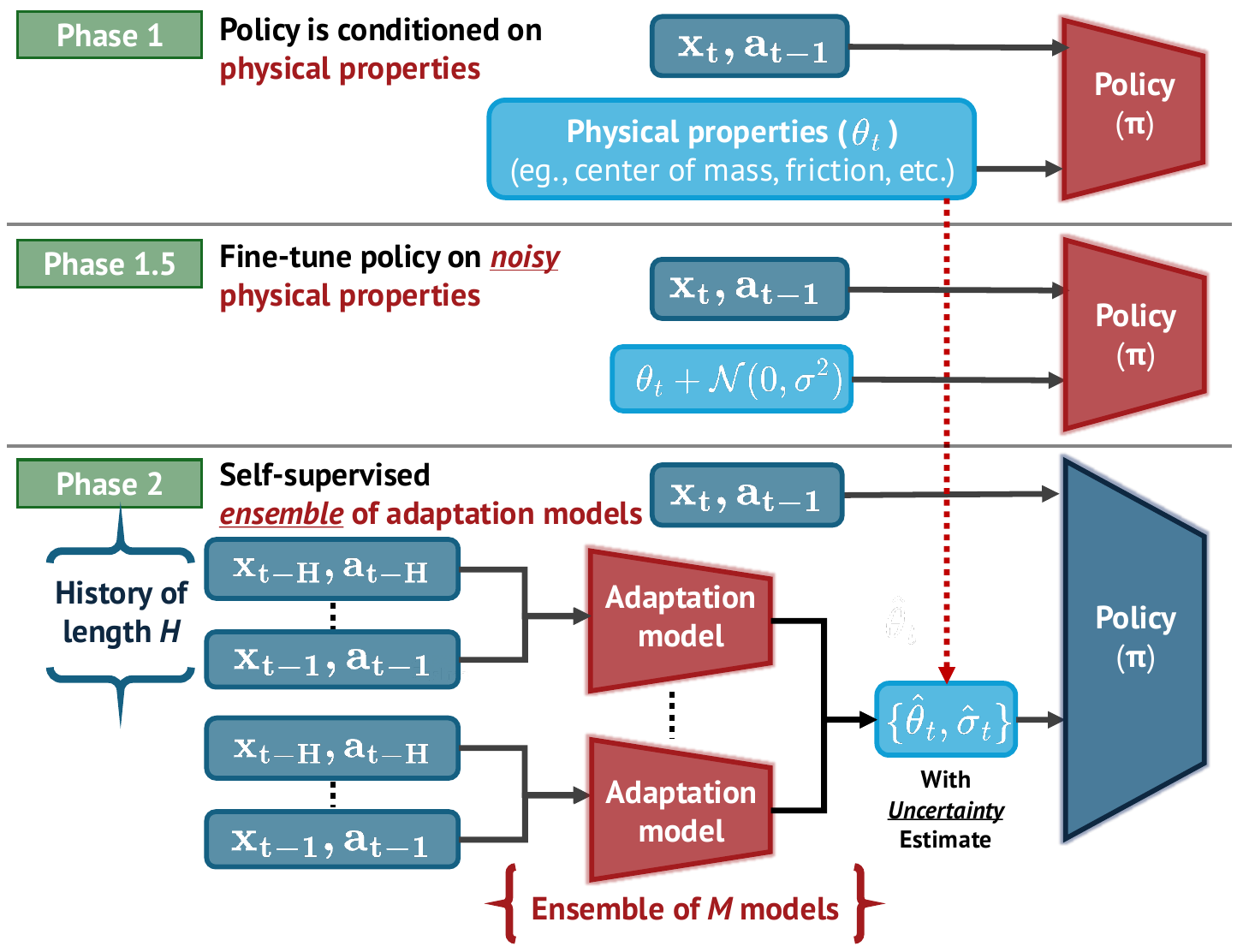}
\caption{\small \textbf{Phys2Real policy training.} The policy and adaptation models are trained in three stages, inspired by RMA \cite{kumar_rma_2021}. \textbf{Phase 1}: the policy is conditioned on ground truth physical properties (e.g., CoM) from simulation, \textbf{Phase 1.5}: fine-tune with \textit{noisy} physical properties to build robustness to downstream noisy estimates from the fused VLM and adaptation estimate. \textbf{Phase 2}: train an ensemble of $M$ encoders that take in a history of observations and actions. The variance of the ensemble estimates provides an \textit{epistemic} uncertainty (Eq.~\ref{eq:epistemic}). Each encoder outputs a physical property estimate and an associated uncertainty, representing the model's \textit{aleatoric} uncertainty (Eq.~\ref{eq:aleatoric}). At test time, we fuse the adaptation estimates and VLM estimates via inverse-variance weighting (Eq.~\ref{eq:fusion}).}
\label{rl-policy}
\end{figure}

\subsection{Physics-Conditioned Policy Learning}

We train deep RL policies in simulation to manipulate objects.
Motivated by evidence that policies with access to physical parameter information can improve performance over domain randomization approaches in locomotion~\cite{kumar_rma_2021}, we explicitly condition the policy networks on physical parameter values.



To enable real-world deployments, we perform a second phase of training, during which the RL policy is frozen, and another neural network adaptation model learns to predict physical properties from historical state-action sequences.  
This approach is inspired by rapid motor adaptation (RMA)~\cite{kumar_rma_2021}, which follows a similar two-phase learning paradigm: in the first phase, a physics encoder uses privileged information available only in simulation to produce a latent $z$ (representing physical properties), which is then input into the RL policy. In the second phase, history information is fed into an adaptation model, which regresses on the latent $z$ trained with the physics encoder from the first phase. 

While RMA produces effective latent estimates in settings like locomotion, general manipulation tasks such as non-prehensile manipulation can often induce uninformative interaction histories, leading to poor environment parameter estimates. We propose to alleviate this by (1) making the physics encoder uncertainty-aware, so that uninformed or poor guesses can be detected, and (2) fusing online information with VLM-informed visual estimates. 

Thus, we extend RMA to \textit{explicitly estimate physical parameters and their uncertainties}. By explicitly estimating physical parameters rather than latent vectors, we obtain interpretable outputs that are in the same representational space as VLM estimates. This is critical so that they can be fused with a linear combination weighted by respective uncertainties as described in Sec.~\ref{sec:method:simtorealtransfer}. 
This VLM prior is important for contact-rich, long-horizon manipulation because object physical properties may be unobservable before or during contact. A reasonable estimate of the object's physical parameters may be crucial to the policy's effectiveness. 

To summarize, our policy training procedure consists of: 
\begin{enumerate}[leftmargin=*]
    \item \textbf{Phase 1}: The policy is trained conditioned on ground-truth physical parameters available in simulation. This enables the policy to learn optimal behaviors for different physical configurations. Unlike standard RMA, we condition directly on interpretable physical parameters rather than learned latents, which is critical for downstream combination with VLM estimates.
    \item \textbf{Phase 1.5 (optional)}: Since the policy at training time observes only ground truth physical parameter values, slightly incorrect VLM or physics encoder estimates at deployment time could lead to out-of-training-distribution states. We thus fine-tune the policy with noisy parameter estimates by applying random noise to each physical parameter during each episode. Our policies were fine-tuned with a Gaussian noise of $\sigma = 1.5\,\mathrm{cm}$.
    
    \item \textbf{Phase 2:} We freeze the learned policy weights and train an ensemble of ${M=10}$ adaptation models that learn to predict physical parameters from a history of observations and actions over a sliding window of ${H=10}$. 
    These models allow the policy to perform online adaptation during deployment. We obtain uncertainty estimates for the physical parameters using ensemble-based uncertainty quantification that captures both epistemic (model uncertainty) and aleatoric (data uncertainty) components. 
\end{enumerate}

As shown in Fig.~\ref{rl-policy}, the actor is conditioned on object pose, robot end-effector position, and object physical properties (e.g., friction, CoM). The critic receives privileged observations, including object velocity and pose. 
We train policies using proximal policy optimization (PPO)~\cite{schulman_proximal_2017} with 4096 parallel environments and an asymmetric actor-critic architecture in IsaacLab~\cite{mittal_orbit_2023}.

\subsection{Sim-to-Real Transfer with Physical Parameter Estimation}
\label{sec:method:simtorealtransfer}

To combat uninformative interaction histories, Phys2Real combines physical parameter estimates from online interaction with VLM-based visual reasoning. 
However, it is unclear how to effectively combine this information. One na\"ive approach is to simply condition the policy or adaptation model on the VLM estimate. But, this introduces additional training time hyperparameters, effectively increasing the state space. Instead, we use the insight that these estimates can be treated effectively as sensing information that can be fused at test time based on their \textit{relative uncertainty}. When the uncertainty of the adaptation model is high, increased reliance can be placed on the VLM estimate, and vice versa.

To instantiate this, both the VLM and adaptation model ensembles provide an estimate of the physical property and an associated uncertainty. We then fuse these estimates using inverse-variance weighting.

\begin{figure}[t]
\centering
\includegraphics[width=0.5\textwidth]{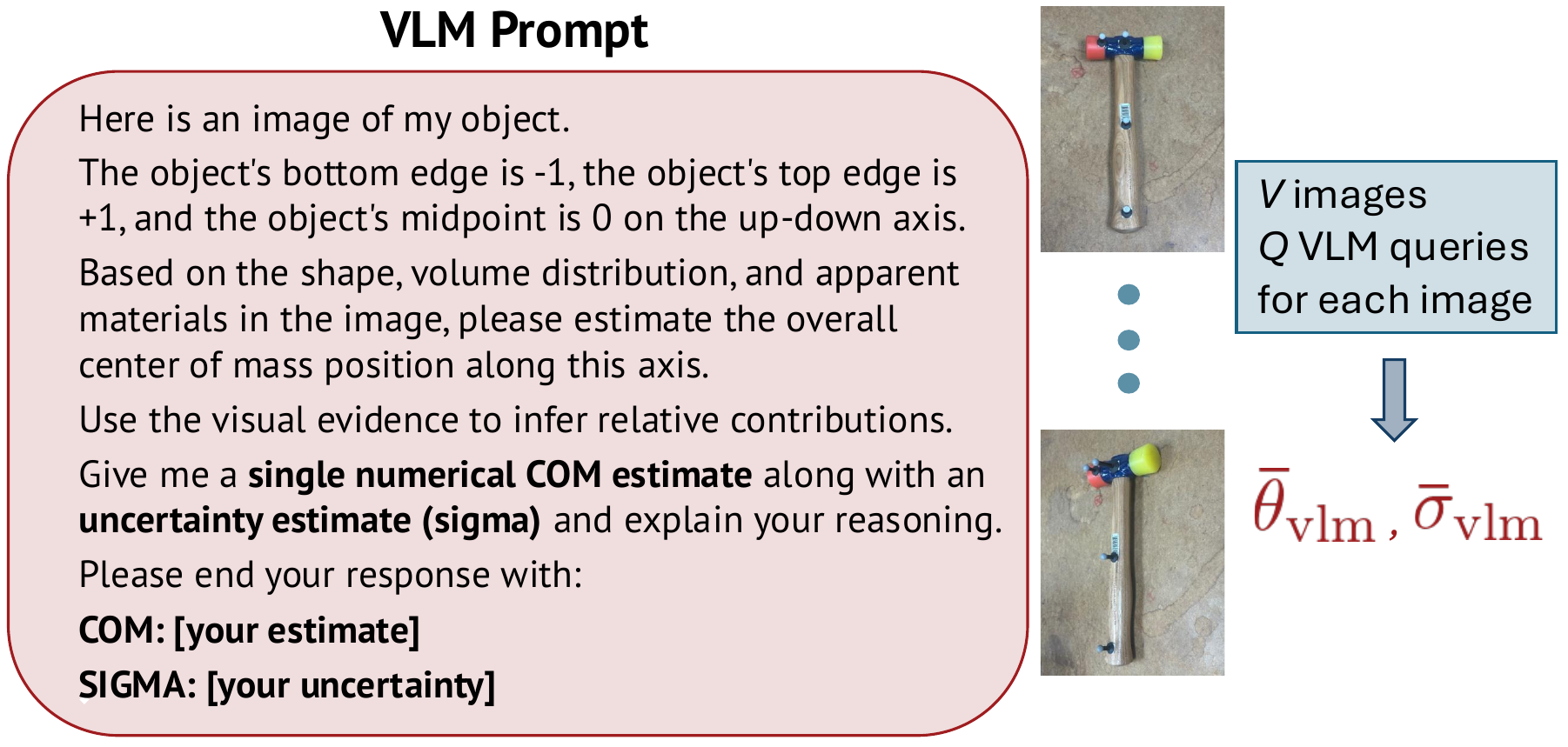}
\caption{\small \textbf{VLM priors for task-relevant physical parameters.} We query a VLM (GPT-5 \cite{openai2025gpt5}) to provide an estimated CoM and uncertainty given an image of the object. For each of $V$ images, we repeat the query $Q$ times. We then calculate the average VLM estimate $\bar{\theta}_{\textrm{vlm}}$ along with the average uncertainty $\bar{\sigma}_{\textrm{vlm}}$. This is fused with the RMA estimate using Eq.~(\ref{eq:fusion}).}
\label{vlm-prompts}
\end{figure}

\subsubsection{Obtaining \texorpdfstring{$\theta_{\mathrm{vlm}}$ and $\sigma_{\mathrm{vlm}}$}{mean and standard deviation of the VLM-inferred physical parameter}}

We query GPT-5~\cite{openai2025gpt5} to estimate task-relevant physical parameters from images. For both the T-block and the hammer, we estimate the CoM along the vertical stem, which affects its rotational dynamics. As shown in Fig. \ref{vlm-prompts}, we capture $V$ images from different viewpoints and query the VLM $Q$ times for each image. We use the prompt shown in Fig.~\ref{vlm-prompts} to obtain physical parameter and uncertainty estimates. 

Example VLM responses and estimation biases are provided in Appendix~\ref{appendix:vlm-com-estimation-exs}.

The VLM estimate is computed as the aggregate mean across the prompts:
    $\theta_{\textrm{vlm}} = \frac{1}{V \times Q}{\sum_{i=1}^V \sum_{j=1}^{Q} \theta_{i, j}}$.

To calculate $\sigma_{\mathrm{vlm}}$, we take the mean of the uncertainty estimates from the VLM. We find empirically that computing the standard deviations of the estimates themselves can cause poor uncertainty estimates (i.e., the VLM may be confidently wrong). However, the model's estimate of its own uncertainty takes into account the uncertainty of an object's properties based purely on visual appearance.

\begin{table*}[b]
\centering
\small
\vspace{2ex}
\caption{\small T-block pushing with the weight at the \textbf{top} (placed $9.5\,\mathrm{cm}$ above the geometric center along the vertical axis; 21 trials per method with initial orientations varied every $45^{\circ}$).
Arrows indicate whether higher ($\uparrow$) or lower ($\downarrow$) values are better, with the top-two values shown in bold.
The privileged baseline is shaded in orange, and our method (Phys2Real) is highlighted in green with a star (*).
}
\label{tab:tblock-top}
\resizebox{\textwidth}{!}{%
\begin{tabular}{ l |c c c c}
\toprule
\textbf{Method} &
\textbf{Success Rate (\%, $\uparrow$)} &
\textbf{Pos. Err $\pm$ Std (cm, $\downarrow$)} &
\textbf{Orient. Err $\pm$ Std (deg, $\downarrow$)} &
\textbf{Time (s, $\downarrow$)} \\
\midrule
\rowcolor{green!10}{\bfseries\boldmath Phys2Real* [$\mathrm{CoM}=+4.0\,\mathrm{cm}$, $\boldsymbol{\sigma}=1.4\,\mathrm{cm}$]} &
\textbf{57.14} & \textbf{2.60 $\pm$ 0.90} & \textbf{2.62 $\pm$ 1.73} & 39.58 \\

Physics-conditioned [VLM=$+4.0\,\mathrm{cm}$] &
4.76 & 10.10 $\pm$ 14.33 & 12.56 $\pm$ 32.54 & 40.80 \\

\rowcolor{orange!10}Physics-conditioned [privileged=$+6.1\,\mathrm{cm}$] &
\textbf{90.48} & \textbf{1.90 $\pm$ 0.98} & \textbf{1.78 $\pm$ 1.71} & 43.43 \\

RMA-only [adaptation model]~\cite{kumar_rma_2021} &
14.29 & 7.60 $\pm$ 8.10 & 4.70 $\pm$ 4.44 & \textbf{37.23} \\

DR [$-3.5\,\mathrm{cm}$, $+7.5\,\mathrm{cm}$] &
23.81 & 6.00 $\pm$ 5.78 & 6.77 $\pm$ 6.90 & \textbf{37.00} \\

Diffusion Policy (DP)~\cite{chi_diffusion_2024} &
20.83 & 25.90 $\pm$ 16.89 & 57.38 $\pm$ 62.50 & 54.14 \\
\bottomrule
\end{tabular}
}
\end{table*}

\begin{table*}[b]
\centering
\small
\caption{\small T-block pushing with the weight at the \textbf{bottom} (placed $6.5\,\mathrm{cm}$ below the geometric center along the vertical axis; 24 trials per method with initial orientations varied every $45^{\circ}$).}
\label{tab:tblock-bottom}
\resizebox{\textwidth}{!}{%
\begin{tabular}{ l |c c c c}
\toprule
\textbf{Method} &
\textbf{Success Rate (\%)} &
\textbf{Pos. Err $\pm$ Std (cm)} &
\textbf{Orient. Err $\pm$ Std (deg)} &
\textbf{Time (s)} \\
\midrule
\rowcolor{green!10}{\bfseries\boldmath Phys2Real* [$\mathrm{CoM}=+0.76\,\mathrm{cm}$, $\boldsymbol{\sigma}=1.55\,\mathrm{cm}$]} &
\textbf{100.00} & \textbf{1.76 $\pm$ 0.54} & 4.73 $\pm$ 2.68 & 44.28 \\

Physics-conditioned [VLM=$+0.76\,\mathrm{cm}$] &
91.67 & \textbf{1.90 $\pm$ 0.59} & \textbf{3.18 $\pm$ 1.69} & 38.27 \\

\rowcolor{orange!10}Physics-conditioned [privileged=$-0.71\,\mathrm{cm}$] &
\textbf{95.83} & 1.92 $\pm$ 0.50 & \textbf{2.76 $\pm$ 1.75} & \textbf{37.80} \\

RMA-only [adaptation model]~\cite{kumar_rma_2021} &
79.17 & 2.23 $\pm$ 0.81 & 3.62 $\pm$ 2.10 & 45.22 \\

DR [$-3.5\,\mathrm{cm}$, $+7.5\,\mathrm{cm}$] &
79.17 & 7.14 $\pm$ 11.34 & 11.11 $\pm$ 13.86 & \textbf{37.50} \\

Diffusion Policy (DP)~\cite{chi_diffusion_2024} &
50.00 & 19.34 $\pm$ 22.20 & 40.71 $\pm$ 52.00 & 38.71 \\
\bottomrule
\end{tabular}
}
\end{table*}

\subsubsection{Obtaining \texorpdfstring{$\theta_{\mathrm{rma}}$ and $\sigma_{\mathrm{rma}}$}{mean and standard deviation of the RMA-inferred physical parameter}}
Given a sequence of observations \{$o_t, o_{t-1}, \dots, o_{t-k}$\} (where each observation includes the previous action, $a_{t-1}$), each ensemble member $i$ produces an estimate $\theta_i$ of the physical parameter (e.g., CoM). The ensemble mean provides our parameter estimate,
%
and the ensemble variance, given by
\begin{equation}
    \sigma^2_{\textrm{epistemic}} = \frac{1}{M} \sum (\theta_i - \theta_{\mathrm{rma}})^2.
    \label{eq:epistemic}
\end{equation}
is the epistemic uncertainty. 
Each ensemble member also estimates aleatoric uncertainty by outputting a mean $\mu_i$ and variance $\sigma^2_i$, trained using a Gaussian negative log-likelihood loss\cite{nix_estimating_1994}. The ensemble's mean aleatoric uncertainty is thus 
\begin{equation}
    \sigma^2_{\mathrm{aleatoric}} = \frac{1}{M} \sum \sigma^2_i.
    \label{eq:aleatoric}
\end{equation}
Our total RMA uncertainty combines both sources:
\begin{equation}
\sigma^2_{\mathrm{rma}} = \sigma^2_{\mathrm{epistemic}} + \sigma^2_{\mathrm{aleatoric}}.
\label{eq:rma}
\end{equation}
This decomposition of uncertainties separates epistemic uncertainty, which captures model disagreement, from aleatoric uncertainty, which captures irreducible noise in observations and the environment. By combining both with a VLM prior,
our method enables robust uncertainty-aware adaptation even when the robot is not in continuous contact with the object. 



\subsubsection{Fusing Estimates With Inverse-Variance Weighting}

We combine physical property estimates from a VLM with estimates from RMA using uncertainty-based weights.
 
This fusion mechanism allows the system to assign greater weight to VLM physical property estimates when the RMA interaction history is uncertain, and conversely, rely more on RMA when the VLM estimates are less certain. The physically interpretable nature of $\theta$ enables direct combination of VLM priors with online adaptation, unlike approaches that use learned latent representations.

The fused estimate $\hat{\theta}$ is computed using inverse-variance weighting:
\begin{equation}
\hat{\theta} =
\frac{\theta_{\mathrm{vlm}}/\sigma_{\mathrm{vlm}}^{2} + \theta_{\mathrm{rma}}/\sigma_{\mathrm{rma}}^{2}}
{1/\sigma_{\mathrm{vlm}}^{2} + 1/\sigma_{\mathrm{rma}}^{2}},
\label{eq:fusion}
\end{equation}
where $\theta_{\mathrm{vlm}}$ and $\theta_{\mathrm{rma}}$ are estimates from the VLM and the ensemble of adaptation models, and $\sigma_{\mathrm{vlm}}$ and $\sigma_{\mathrm{rma}}$ are their respective uncertainties. $\hat{\theta}$ is used to condition the policy at test time. Under the assumption that the estimates are unbiased with independent noise and known variances, this estimator is theoretically justified as the Best Linear Unbiased Estimator (BLUE), which minimizes the variance of the fused estimate.

\begin{figure*}[!tbp] 
\centering

\subfloat[{\bfseries\boldmath CDF for weight at top configuration for position errors $\leq 10\,\mathrm{cm}$}]{
    \includegraphics[width=0.48\textwidth]{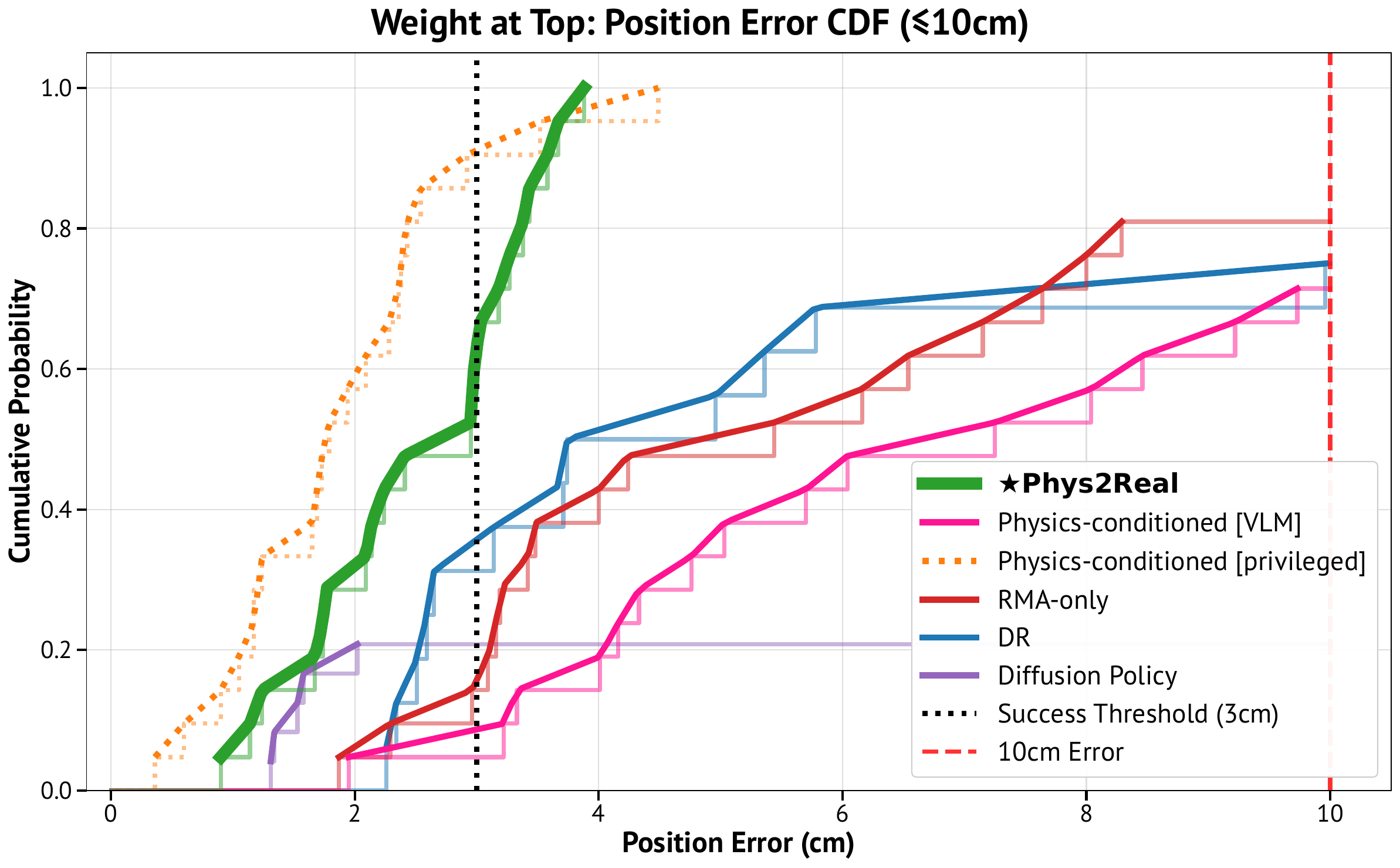}
    \label{fig:tblock-weighttop}
}
\subfloat[{\bfseries\boldmath CDF for weight at bottom configuration for position errors $\leq 5\,\mathrm{cm}$}]{
    \includegraphics[width=0.48\textwidth]{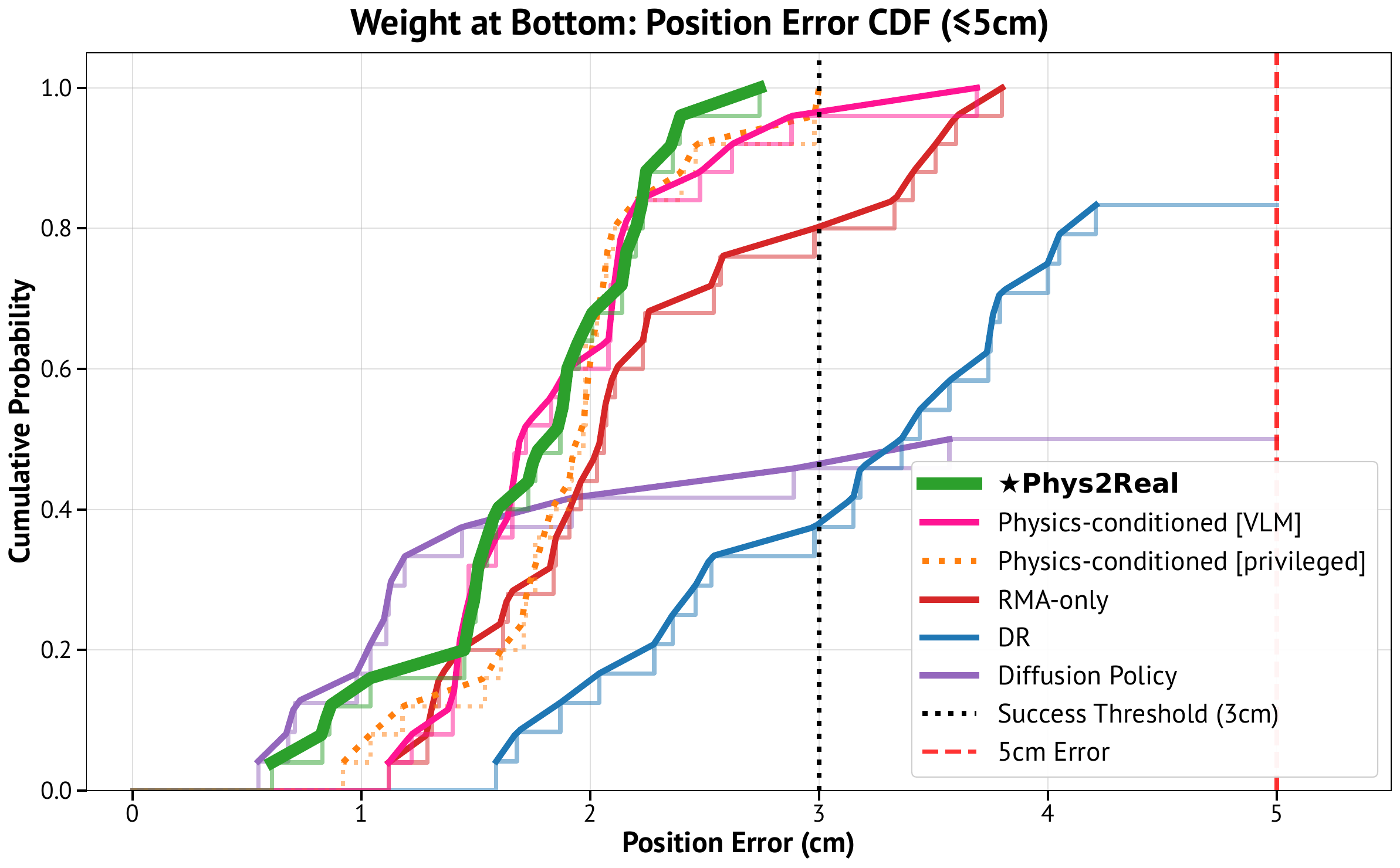}
    \label{fig:tblock-weightbottom}
}

\caption{
\small \textbf{T-block pushing task results.} We compare the cumulative distribution functions (CDFs) of position errors at the end of rollouts from each policy, which show the percentage of experimental trials (y-axis) with position error less than or equal to a given value (x-axis). For an optimal policy, the curve should rise steeply and stay toward the left, indicating that most trials have low final error. We evaluate T-block pushing under two weight configurations:
(a) weight at the top and (b) weight at the bottom. Overall, Phys2Real (green) consistently has low position error throughout the percentiles.
}
\label{fig:tblock-comparison}
\end{figure*}

\section{Experiments} \label{experiments}

In our experiments, we aim to understand how Phys2Real compares against baseline and privileged methods by investigating the following research questions:

\noindent\textbf{Q1:} Can accurate physical parameter estimation \textit{improve policy performance} for robotic manipulation? 
How does Phys2Real compare with a physics-conditioned sim-to-real RL policy with privileged ground truth physical properties?

\noindent\textbf{Q2:} Are task-relevant physical parameters  \textit{estimated by VLMs} sufficient to improve the performance of a physics-conditioned sim-to-real RL policy?

\noindent\textbf{Q3:} Are task-relevant physical parameters \textit{estimated through interaction adaptation} sufficient to improve the performance of a physics-conditioned sim-to-real RL policy?

\noindent\textbf{Q4:} Can Phys2Real improve policy performance when applied to \textit{real-world objects without available object models}? 

To address these questions, we evaluate Phys2Real on two non-prehensile manipulation tasks that require accurate physical modeling for successful execution:
\begin{itemize}[leftmargin=*]
    \item \textbf{T-block pushing:} To answer (\textbf{Q1})-(\textbf{Q3}), we vary the CoM along the vertical axis of the T-block by adding a small metal weight of 143 grams. We evaluate all policies using two different configurations based on where the metal weight is placed, which alters the CoM of the object along the vertical axis: (1) \textbf{weight on the top}, where the weight is placed at $9.5\,\mathrm{cm}$ above the geometric center of the T-block, and (2) \textbf{weight on the bottom}, where the weight is placed $6.5\,\mathrm{cm}$ below the geometric center. This makes the ground truth CoM $6.1\,\mathrm{cm}$ and $-0.7\,\mathrm{cm}$ for the weights at top and bottom configurations, respectively.
    This task tests adaptation to varying object CoMs, which impacts the block's rotational dynamics during pushing. We use a known T-block mesh (as in DP~\cite{chi_diffusion_2024}) to isolate the effects of physical parameter estimation. 
    \item \textbf{Hammer pushing:} The CoM near the head of the hammer creates complex motion dynamics that must be accounted for during manipulation. We generate the hammer mesh through our real-to-sim reconstruction pipeline described by Fig.~\ref{mesh-reconstruction}. We answer (\textbf{Q4}) with this task.
\end{itemize}

For each task, we measure four metrics: (1) success rate, (2) final position error in meters, (3) final orientation error in degrees, and (4) task completion time in seconds. We define success as achieving less than $3\,\mathrm{cm}$ positional error and less than $20^{\circ}$ orientation error relative to the target pose. 

\subsection{Experimental Setup}

We experiment using a 6-DOF UFactory xArm robotic arm. The robot uses a cylindrical end-effector to push objects on a table. Its observations consist of object pose, robot end-effector $xy$ positions, and estimated physical parameters. Its actions are changes in end-effector $xy$ positions. 

For real-world evaluation, we use motion capture to obtain accurate object poses. However, Phys2Real is designed to also handle visual inputs alone, and replacing motion capture with perception-based tracking is an exciting future direction.

\subsection{Real-world Evaluation: T-block Pushing}

Tables~\ref{tab:tblock-top}~and~\ref{tab:tblock-bottom} show the performance of Phys2Real compared to several baselines under the two conditions (weight at top and bottom), which shift the object's CoM and result in distinct pushing dynamics and levels of difficulty for sim-to-real transfer. We also compare the performance of each method from the perspective of the cumulative distribution functions (CDF) of position errors in Figs. \ref{fig:tblock-weighttop} and \ref{fig:tblock-weightbottom}. In these plots, the optimal policy has low error at every percentile. 

\subsubsection{Effect of Physical Parameter Estimation}
Firstly, we would like to understand whether having an accurate estimate of the object's physical properties is beneficial for manipulation. To address \textbf{Q1}, we first compare Phys2Real against the following three methods.

The first two methods are baselines, DR and diffusion policy (DP), that are physics parameter \textit{unaware}. For DR, the location of the weight on the T-block is randomized in simulation from the bottom to the top of the block. We use DP~\cite{chi_diffusion_2024} as a strong imitation learning baseline that is also unaware of object physics parameters. 
We train a state-based DP using motion capture data, matching our state-based RL policies. 
We collect $100$ demonstrations, varying the weight's location on the vertical axis of the T-block and the T-block's initial orientation in each episode, which is comparable to the domain randomization in simulation. 
We follow the implementation of ~\cite{chi_diffusion_2024}.


\begin{table*}[!t]
\centering
\small
\vspace{1mm}
\caption{\small \textbf{Hammer pushing results.} Both Phys2Real and DR achieve 100\% success rates across 15 trials.
Here, the hammer's center of mass (CoM) and uncertainty ($\sigma$) are provided by the VLM: $\mathrm{CoM}=8.90\,\mathrm{cm}$, $\sigma = 1.2\,\mathrm{cm}$.
To prevent the robot from continually pursuing small improvements, a trial is deemed successful if the hammer satisfies the success criteria for at least 2s.}
\label{tab:hammer-results}
\resizebox{\textwidth}{!}{%
\begin{tabular}{ l |c c c c}
\toprule
\textbf{Method} &
\textbf{Success (\%, $\uparrow$)} &
\textbf{Pos. Err (cm, $\downarrow$)} &
\textbf{Orient. Err (deg, $\downarrow$)} &
\textbf{Time $\pm$ Std (s, $\downarrow$)} \\
\midrule
\rowcolor{green!10}{\bfseries\boldmath Phys2Real* [$\mathrm{CoM}=+8.9\,\mathrm{cm}$, $\boldsymbol{\sigma}=1.2\,\mathrm{cm}$]} &
\textbf{100.00} & \textbf{1.74} & \textbf{4.68} & \textbf{77.79 $\pm$ 44.08} \\
DR [$-13\,\mathrm{cm}$, $+13\,\mathrm{cm}$] &
\textbf{100.00} & \textbf{1.55} & \textbf{2.89} & \textbf{90.65 $\pm$ 42.03} \\
\bottomrule
\end{tabular}
}
\end{table*}


The third method is a physics-conditioned policy with \textit{privileged} information on the ground truth physics parameter, computed from the known T-block geometry assuming uniform density and the metal weight's measured mass. This baseline represents the oracle upper bound on performance.


When the weight is placed at the bottom of the T-block (Table~\ref{tab:tblock-bottom}), Phys2Real achieves 100\% success and the lowest position error across all methods. In comparison, DR and RMA without a VLM prior show lower success rates and higher errors. Diffusion policy also performs poorly. The results indicate that Phys2Real can match the performance of a privileged oracle by fusing vision-based priors and interactive adaptation, without using privileged information.

When we turn to the position error CDF in (Fig. \ref{fig:tblock-weightbottom}) for a more fine-grained understanding of position errors, we see that the Phys2Real curve rises sharply and up, indicating consistently low position error across nearly all trials. While DP has low error for around 40 percent of trials, the long tail is high. DR similarly has a long tail of high errors, showing that DP and DR easily become out of distribution, leading to high position errors.
When the weight is at the top of the block, shown in Table~\ref{tab:tblock-top}, all policies struggle to achieve a high success rate, except the privileged CoM-conditioned policy with a 90.48\% success rate. We hypothesize that the higher CoM causes more unpredictable dynamics. Even in this case, Phys2Real is the second-best performer at 57.14\%.



\begin{figure}[!htbp]
  \centering

\subfloat[\textbf{Object and end-effector trajectories.} The end-effector (blue) and T-block (orange) paths are shown as the block moves $40\,\mathrm{cm}$ to the right from an initial orientation of $180^{\circ}$ (light blue outline) to a final orientation of $0^{\circ}$ (dark blue filled shape) at the target location (gold star). The robot achieves a final position error of $1.7\,\mathrm{cm}$.]{%
  \includegraphics[width=\linewidth]{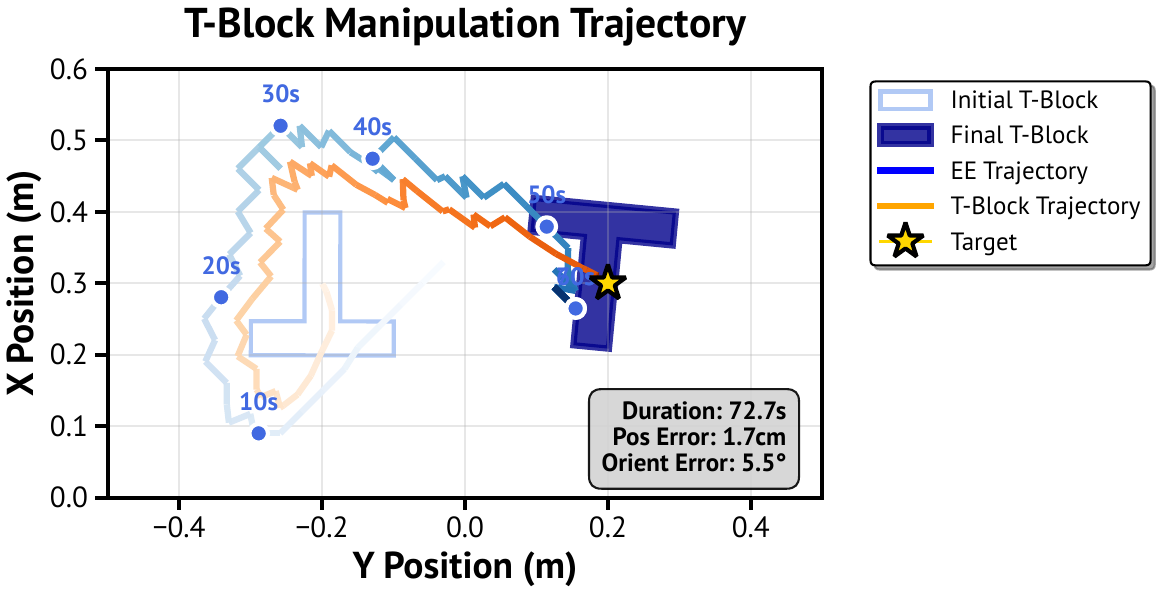}%
  \label{fig:tblock_trajectory}

}


\subfloat[\textbf{Parameter estimates over time.} The VLM prior (red) provides an initial estimate ($\theta_{\mathrm{vlm}} = 4.0\,\mathrm{cm}$, $\sigma_{\mathrm{vlm}} = 1.4\,\mathrm{cm}$).]{%
  \includegraphics[width=\linewidth]{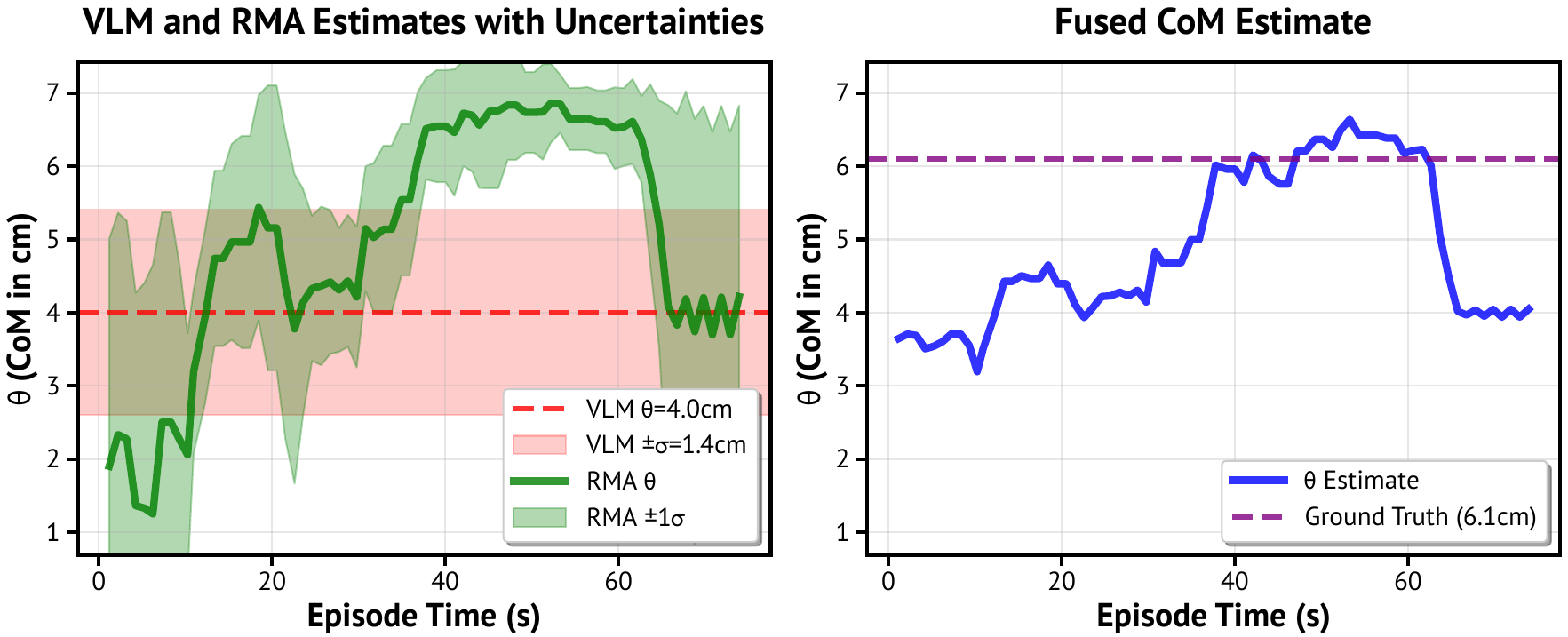}%
  \label{fig:tblock_timeseries}
}

%

  \caption{\small \textbf{T-block with weight at the top.} 
  Early in the task, the RMA estimate (green) is highly uncertain and far from the ground truth (${\sim}6.0\,\mathrm{cm}$). 
  As contact continues, uncertainty decreases and the fused estimate (blue, Eq.~\ref{eq:fusion}) converges toward the ground truth at around 40s. 
  After 55s, when contact ends, uncertainty rises again due to lack of new information.}
  \label{fig:tblock_complete}
\end{figure}

\subsubsection{Impact of VLM-Only or RMA-Only Physical Parameter Estimation on Policy Performance}


For \textbf{Q2}, we perform an ablation without the adaptation model, conditioning the policy only on the VLM estimate. Conversely, for \textbf{Q3}, we remove the VLM estimate and only condition the policy on the adaptation model's estimate.

Before evaluating these ablations, we first visualize how the fused estimate evolves. Fig. \ref{fig:tblock_trajectory} shows Phys2Real in a representative rollout with weight at the top, where the initial CoM estimate is uncertain (shown in Fig. \ref{fig:tblock_timeseries}). As the robot interacts with the object, the uncertainty (in green) of the adaptation models decreases. The fused estimate (blue) converges to the ground truth of $6\,\mathrm{cm}$. After interaction ends (60\,s), uncertainty rises due to the lack of information about the CoM. This example illustrates how the VLM prior and interactive adaptation models are fused. We next evaluate how removing either source affects performance.

In the challenging weight at the top case, Phys2Real (57.14\%) significantly outperforms the policy conditioned on just the VLM estimate (4.7\%) as well as RMA-only (14.29\%). 
This showcases that \textbf{neither the VLM estimate nor the interactive adaptation is enough to perform well individually; both sources of information are necessary for the policy to succeed}. We hypothesize that this is because the VLM CoM estimate is around $2\,\mathrm{cm}$ below the ground truth CoM, so the physics-conditioned on the VLM estimate policy pushes the object at more out-of-distribution locations. Similarly, without a VLM prior, the RMA-only policy pushes the object to out-of-distribution locations; because the initial parameter estimate is near the mean of the CoM distribution during Phase 2 training, it takes time for the adaptation model to adapt, and the first few initial pushes create out-of-distribution scenarios.

When the weight is placed on top (Fig. \ref{fig:tblock-weighttop}), the differences become more pronounced. The privileged RMA phase 1 policy has the steepest CDF near the low-error region, while
Phys2Real tracks closely behind, showing its ability to handle challenging, unstable CoM dynamics without privileged information. RMA-only and VLM-only baselines diverge further to the right,
reflecting large errors and high failure rates. DP exhibits the longest tails. We hypothesize this is because during data collection, the human demonstrator can observe the weight location, which biases their actions, but this information is absent at test time.

These results show that Phys2Real consistently shifts both average performance and the full error distribution toward lower values, outperforming baselines and approaching the performance of a policy with privileged ground truth CoM information across both easy and challenging configurations. 



We provide additional ablations evaluating robustness to different VLM priors in Appendix~\ref{appendix:phase15}.



\subsection{Real-world Evaluation: Hammer Pushing}


Now that we have shown the success of Phys2Real on the T-block with a known mesh, can Phys2Real be extended to objects encountered in the real world (\textbf{Q4})? To investigate this question, we transform real-world objects into simulation-ready assets. As shown in Fig.~\ref{mesh-reconstruction}, images of the hammer are segmented with SAM-2~\cite{ravi_sam_2024} and reconstructed into object-centric GSplats using SuGaR~\cite{guedon_sugar_2023}. We mirror the GSplat across its primary axis of symmetry and apply the Marching Cubes algorithm to extract a clean, watertight mesh. While this pipeline works well for approximately symmetric objects like T-blocks and hammers, mirroring can distort the true shape and mass distribution of asymmetric objects. Extending to asymmetric objects would require alternative meshing strategies to preserve geometric fidelity.

For hammer pushing, both Phys2Real and DR achieve 100\% success across 15 trials, as shown in Table \ref{tab:hammer-results}, demonstrating that the task is solvable for both methods with sufficient training. However, Phys2Real achieves faster task completion, with an average time of 77.79\,s vs. 90.65\,s for DR, a 14.2\% improvement in task completion time, indicating that Phys2Real produces more efficient trajectories. Final position and orientation errors are comparable between the two methods.

\section{Future Work}

Our work presents many areas of future work, including the following:

\noindent \textbf{Uncertainty-aware multi-parameter estimation.} 
Our work focuses on estimating the center of mass along a single axis. Real-world manipulation requires reasoning about multiple interacting physical properties, including friction, mass, and stiffness. While our framework is designed to accommodate additional physical parameters, the practical challenges of scaling to more physical parameters are left to future work.
Furthermore, to account for estimation uncertainty through $\sigma_{\mathrm{vlm}}$, we take the mean of the VLM's self-reported uncertainty estimates. However, several principled approaches could be used instead. For instance, conformal prediction or next-token prediction probabilities from a multiple-choice formulation (e.g., having the VLM select between discretized CoM ranges) could be explored in future work for better uncertainty calibration. 

\noindent \textbf{Broader object and task generalization.} We evaluate on two objects (T-block and hammer) with planar pushing. Future work could evaluate on a wider range of object geometries and physical properties, including asymmetrical, deformable, and articulated objects. 
Additionally, our results demonstrate that the VLM prior is particularly valuable when interaction histories are intermittent, such as in planar pushing, a non-prehensile task with intermittent contact. A future direction is to explore tasks across the contact spectrum: in contact-rich tasks such as tool use or in-hand manipulation, where the adaptation module receives dense signals and the VLM prior may be less critical, and in impulsive tasks such as throwing or striking, where the outcome is determined by a single contact, making the VLM prior more critical.


Overall, Phys2Real demonstrates that foundation models such as VLMs contain useful physical priors that can be refined through interaction. Extending this idea beyond physical parameter estimation, such as world models for vision-language-action (VLA) models, is a promising direction for building robotic systems that combine the semantic knowledge of foundation models with the physical grounding from interacting with the real world.


\section{Conclusion}

We present Phys2Real, a real-to-sim-to-real pipeline that improves robot manipulation by combining VLM-based physical parameter estimates with interaction-based adaptation through uncertainty-aware fusion. This represents a different paradigm from prior work that uses VLMs primarily for high-level reasoning, and from adaptive control techniques that lack foundation model priors.
In real-world experiments, we show that Phys2Real outperforms DR across multiple tasks and physical configurations, achieving notable performance gains without ground-truth physical information at test time.
This work demonstrates the potential of integrating visual geometry, physical understanding, and adaptive control for robotic manipulation. The shift toward adaptation through vision and interaction opens new directions that leverage foundation model capabilities while grounding them in physical interactions, enabling more adaptive and general robotic systems for novel object manipulation. 
%

\section*{Acknowledgments}

The authors thank John Tucker and Rohan Thakker for valuable discussions and feedback. 

\bibliographystyle{IEEEtran}
\bibliography{IEEEexample}

\vspace{-1mm}

\appendices
\section{Phase~1.5 Ablations}
\label{appendix:phase15}


To further validate the robustness of our uncertainty-aware fusion approach, we investigate the question 
\mbox{\textbf{Q5:}~How robust is Phys2Real to inaccurate} VLM parameter estimates compared to Phase 1.5 training?

Table~\ref{tab:com-variation} shows a comparison between Phys2Real against Phase 1.5 (physics-conditioned policy trained with noisy parameter estimates). For this ablation study, both the Phase 1.5 baseline and the Phase 1.5 training used in Phys2Real's adaptation module use $\sigma = 1.0\,\mathrm{cm}$ Gaussian noise. 

We find that Phys2Real is \textit{robust} to inaccurate CoM estimates, which is evident in its relatively constant success rate (8-10 successes/12 trials). In contrast, the Phase 1.5 policy suffers significant performance degradation as the VLM parameter estimates become less accurate.

As the CoM estimate from the VLM deviates further from the ground truth (from $6.0\,\mathrm{cm}$ to $-2.0\,\mathrm{cm}$), Phys2Real maintains more consistent performance than Phase 1.5. Notably, when ${\mathrm{CoM}=-2.0\,\mathrm{cm}}$, Phys2Real achieves an 83.3\% success rate, compared to only 33.3\% for Phase 1.5.

\vspace{2mm}

\begin{table}[!h]
\centering
\small
\caption{\small
Extended T-block pushing results across varying VLM CoM configurations (12 trials per condition, with initial orientations varied every $90^{\circ}$) for the weight-at-top case.
Phys2Real consistently outperforms the Phase~1.5 baseline, particularly as parameter estimates become less accurate.
}
\label{tab:com-variation}
\resizebox{\columnwidth}{!}{%
\begin{tabular}{l c c c c}
\toprule
\textbf{Method} &
\textbf{Success} &
\textbf{Pos. Err} &
\textbf{Orient. Err} &
\textbf{Time} \\
 & \textbf{(\%)} & \textbf{(cm)} & \textbf{(deg)} & \textbf{(s)} \\
\midrule
\multicolumn{5}{c}{\textit{$\mathrm{CoM}=+6.0\,\mathrm{cm}$, $\sigma = 1.0\,\mathrm{cm}$ (GT $\mathrm{CoM}\approx +6.0\,\mathrm{cm}$)}} \\
\midrule
\rowcolor{green!10}Phys2Real & \textbf{75.0} & 3.7 & 3.9 & \textbf{43.5} \\
Phase 1.5 & \textbf{75.0} & \textbf{2.2} & \textbf{2.0} & 43.9 \\
\midrule
\multicolumn{5}{c}{\textit{$\mathrm{CoM}=+4.0\,\mathrm{cm}$, $\sigma = 2.0\,\mathrm{cm}$}} \\
\midrule
\rowcolor{green!10}Phys2Real & \textbf{83.3} & \textbf{2.2} & \textbf{2.6} & 46.9 \\
Phase 1.5 & 75.0 & 5.1 & 6.5 & \textbf{42.5} \\
\midrule
\multicolumn{5}{c}{\textit{$\mathrm{CoM}=+2.0\,\mathrm{cm}$, $\sigma = 4.0\,\mathrm{cm}$}} \\
\midrule
\rowcolor{green!10}Phys2Real & 66.7 & \textbf{2.2} & \textbf{1.4} & 46.4 \\
Phase 1.5 & \textbf{75.0} & 2.6 & 2.3 & \textbf{43.1} \\
\midrule
\multicolumn{5}{c}{\textit{$\mathrm{CoM}=0.0\,\mathrm{cm}$, $\sigma = 6.0\,\mathrm{cm}$}} \\
\midrule
\rowcolor{green!10}Phys2Real & \textbf{83.3} & \textbf{3.7} & 4.8 & \textbf{41.7} \\
Phase 1.5 & 58.3 & 4.4 & \textbf{3.8} & \textbf{41.7} \\
\midrule
\multicolumn{5}{c}{\textit{$\mathrm{CoM}=-2.0\,\mathrm{cm}$, $\sigma = 8.0\,\mathrm{cm}$}} \\
\midrule
\rowcolor{green!10}Phys2Real & \textbf{83.3} & \textbf{2.5} & 2.7 & 44.5 \\
Phase 1.5 & 33.3 & 3.7 & \textbf{2.4} & \textbf{43.2} \\
\bottomrule
\end{tabular}
}
\end{table}

\section{VLM CoM Estimation Examples}
\label{appendix:vlm-com-estimation-exs}

VLM priors alone may not be physically grounded. While they provide reasonable initial estimates, interactive online adaptation can substantially refine these estimates. For instance, for the T-block with a metal weight at the top, the VLM tends to bias its estimate toward the geometric center, leading the estimate of ${\mathrm{CoM}=+4.0\,\mathrm{cm}}$ to be lower than the ground truth ${\mathrm{CoM}=+6.0\,\mathrm{cm}}$. The standard deviation across $V=5$ images and $Q=5$ queries is $0.9\,\mathrm{cm}$. 


\tcbset{breakable}

The following is an example output to the prompt shown in Fig.~\ref{vlm-prompts}:

\begin{tcolorbox}[stanfordstyle, title=GPT-5 Response Example (\textit{Biased}), breakable]
\small
\textbf{Estimate reasoning:}

\textbf{Geometry:} The object is a ``T'' with a long vertical stem and a thick top crossbar. For a uniform-density T of these apparent proportions (crossbar height $\approx$ stem width; stem length $\approx 5\times$ stem width), the area/volume centroid lies above the geometric mid-height. A quick two-rectangle model gives $\sim +0.20$ on the [$-1$, $+1$] scale.

\textbf{Materials:} The green body appears 3D-printed plastic (moderate density, likely not solid). Added masses:
\begin{itemize}[leftmargin=1em]
  \item A steel block at the very top (heavy, near $+1$).
  \item Five small metal pegs: two in the top bar (high), three on the stem including one near the bottom (lower).
\end{itemize}

\textbf{Weighting:} Using plausible masses (plastic body dominating but with $\sim$20--40\% added metal), the top steel block shifts the CoM upward strongly; the three lower pegs pull it down slightly. Simple mass-weighted averaging from these placements yields a CoM around $+0.3$ to $+0.4$ relative to the center.

\textbf{Uncertainty sources:} Unknown plastic infill, exact metal masses, and small ambiguity about whether the very top of the steel block defines $+1$. These yield a roughly $\pm 0.1$--$0.15$ spread.

\textbf{CoM:} $+0.34$ \quad \textbf{Sigma:} $0.12$
\end{tcolorbox}

\vspace{-0.2em}

It is difficult to estimate physical parameters (e.g., density, mass) entirely from images, which contributes to the inaccurate VLM CoM estimate of $+0.34$ (${\mathrm{CoM}=3.4\,\mathrm{cm}}$). The VLM estimates the CoM of the 3D-printed T-block without the metal weight to be $+0.20$ (${\mathrm{CoM}=2.0\,\mathrm{cm}}$), while the actual CoM without the metal weight is ${\mathrm{CoM}=3.4\,\mathrm{cm}}$. This biased estimate, combined with the incorrect weighting assumptions, causes the CoM estimate of the combined T-block and weight to be lower than the ground truth. 

We next show an example response that yields an estimate closer to the ground truth CoM of $6.0\,\mathrm{cm}$. 

\begin{tcolorbox}[stanfordstyle, title=GPT-5 Response (\textit{Closer to Ground Truth}), breakable]
\small
\textbf{Estimate reasoning:}

The green body is a T-shaped 3D-printed plastic piece. Its surface texture and visible infill pattern suggest low effective density (roughly 20--35\% of solid plastic).

For a uniform-density T of this proportion (stem height $\approx 4\times$ its width; crossbar thickness $\approx$ stem width; crossbar span $\approx 5\times$ stem width), the vertical centroid is around $+0.35$ on the [$-1$, $+1$] scale.

A substantial steel block sits at the very top (near $+0.95$). Given steel’s much higher density and the likely low infill of the green part, that block plausibly contributes a mass comparable to or greater than the plastic body.

Small pegs or fasteners appear much lighter and are distributed across the height; they make only a minor correction.

Weighting the baseline $+0.35$ by a top-mass fraction of roughly 40--60\% for the steel block yields a CoM around $+0.6$ to $+0.7$.

\textbf{Final estimate with uncertainty} (reflecting unknown infill and exact steel mass): \\
\textbf{CoM:} $+0.63$ \quad \textbf{$\sigma$:} $0.17$
\end{tcolorbox}

Here, the VLM accurately predicts the CoM of the T-block without the weight to be at approximately $+0.35$ ($3.5\,\mathrm{cm}$) above the geometric center. In addition, the steel block's contribution pulls the estimate higher by a few more centimeters, resulting in a combined CoM estimate for the T-block and weight of $+0.63$ ($6.3\,\mathrm{cm}$), which more closely matches the ground truth. 

These examples illustrate that while VLMs can provide reasonable qualitative reasoning about object geometry and materials, their estimates may sometimes be biased or physically inconsistent. This highlights the necessity of interactive online adaptation in Phys2Real to refine these estimates, improving overall policy performance.

\end{document}